# Combining LiDAR Space Clustering and Convolutional Neural Networks for Pedestrian Detection

Damien Matti[1], Hazım Kemal Ekenel[1,2], Jean-Philippe Thiran[1]
[1]LTS5, EPFL, Lausanne, Switzerland
[2]SiMiT Lab, ITU, Istanbul, Turkey
damien.matti@epfl.ch, hazim.ekenel@epfl.ch, jean-philippe.thiran@epfl.ch

## Abstract

*Pedestrian detection is an important component for safety of autonomous vehicles, as well as for traffic and street surveillance. There are extensive benchmarks on this topic and it has been shown to be a challenging problem when applied on real use-case scenarios. In purely image-based pedestrian detection approaches, the state-of-the-art results have been achieved with convolutional neural networks (CNN) and surprisingly few detection frameworks have been built upon multi-cue approaches. In this work, we develop a new pedestrian detector for autonomous vehicles that exploits LiDAR data, in addition to visual information. In the proposed approach, LiDAR data is utilized to generate region proposals by processing the three dimensional point cloud that it provides. These candidate regions are then further processed by a state-of-the-art CNN classifier that we have fine-tuned for pedestrian detection. We have extensively evaluated the proposed detection process on the KITTI dataset. The experimental results show that the proposed LiDAR space clustering approach provides a very efficient way of generating region proposals leading to higher recall rates and fewer misses for pedestrian detection. This indicates that LiDAR data can provide auxiliary information for CNN-based approaches.*

## 1 Introduction

One major criterion for a wide diffusion of the autonomous vehicle technology is the ability to significantly reduce the number of road accidents, a task that highly depends on the detection of surrounding agents around the vehicle. Two types of sensors are usually exploited on vehicles to cope with this task: cameras and LiDAR (Light Detection And Ranging). The purpose of the latter is to measure an accurate distance between the sensor and a target. In the case of autonomous vehicles, it provides the distance information of the surrounding obstacles according to the vehicle position. The latest LiDAR sensors can generate a dense 3-D point cloud. Using these sensors, two methods arise from the literature to solve the problem of pedestrian detection. The first one uses a LiDAR sensor and focuses on creating a map of agents in motion around the vehicle by cumulating temporal information [19]. The second approach consists of applying computer vision algorithms on the captured images. With the recent development of deep neural networks for image classification, current state-of-the-art performance is achieved by CNNs. There are also a few studies on multimodal detection [2, 13, 15], in which the camera image is combined with additional information such as a disparity map of a stereo-camera or a dense depth map inferred from depth measurements.

In this paper, we also present a multimodal approach that consists of utilizing depth measurements to create image region proposals and a state-of-the-art CNN, called *ResNet*, for visual object detection [11]. The main objective of this study is to show the usability and usefulness of LiDAR data as an additional source of information. We developed a novel framework to generate proposals from depth measurements. Our hypothesis is that using depth data and prior information about the size of the objects, we can reduce the search space in the images by providing candidates and, therefore, speeding up detection algorithms. In addition, we hypothesize that this prior definition of the location and size of the candidate bounding box will also decrease the number of false detections. The algo-rithm is built upon the idea of clustering the 3-D point cloud of the LiDAR. It starts with raw measurements downsampling, followed by the removal of points belonging to the floor plane. Then, a density-based clustering algorithm generates the candidates that are projected on the image space to provide a region of interest. An overview of our method is shown in Figure 1. The proposed approach is evaluated on the KITTI dataset. We have observed that exploiting LiDAR data results in higher recall rates and less misses for pedestrian detection.

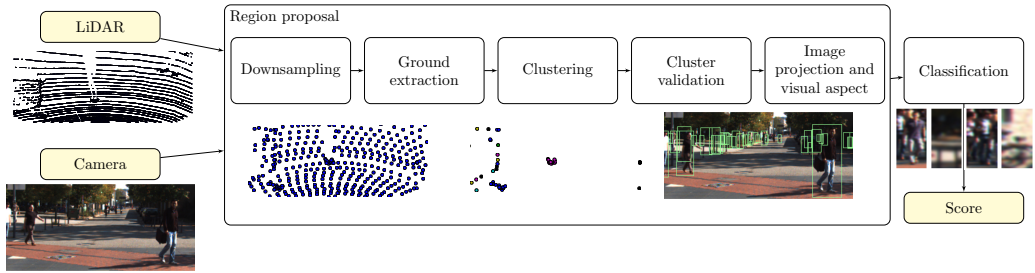

Figure 1: Detection framework

The rest of the paper is organized as follows. In section 2, a summary of previous work on related topics is given. Section 3 describes the methodology used in this work. Section 4 presents relevant metrics to quantify the efficiency of our approach and the experimental results. Finally, conclusions and future work are summarized in section 5.

## 2 Related work

In the next subsections, we will briefly present LiDAR-based detection and visual pedestrian detection methods. A comprehensive overview of previous pedestrian detection approaches can be found in [1].

### 2.1 LiDAR-based Detection

LiDAR sensors have been popular since the birth of autonomous vehicles. Different approaches exploit these sensors. The most common is the creation of an occupancy grid map. This map represents the laser measurement density and is generated by computing a two dimensional histogram of the point cloud that has been projected on the x-y plane. A probability estimate of the existence of an obstacle is then evaluated by computing the posterior probability based on temporal data. This Bayesian temporal filtering emphasizes the surrounding objects in motion [19]. The occupancy grid map allows to identify static objects from moving objects. It has been intensively used for detecting the surroundings of a vehicle in order to monitor and predict the movement of other road users. Another approach is to use a four-layer laser to detect pedestrians by filtering the signal of each laser plane separately and performing a fusion of the different detections [9]. Among LiDAR-camera fusion schemes, other researchers are using a *depth map* along with color channels to perform the detection. The *depth map* is a "dense" representation of LiDAR measurements. Premebida *et al.* [15] proposed to use "sparse" laser data to generate a dense *depth map* of the size of the image using bilateral filtering. The same idea can be applied in a visual approach by using a stereo camera to generate a *disparity map* that replaces the *depth map* [2, 13].

Our method shares similar ideas as in [2, 5, 7, 12, 14]: using the depth information to reduce the search space in the image. In [5], the authors describe a car detection and tracking algorithm based on a single layer LiDAR. They first cluster the LiDAR data before reconstructing the original shape of each object based on temporal information. Spinello *et al.* [18] propose a comparable approach: a one-layer laser range is used to cluster and classify the objects. In parallel, the clusters are also classified in the image space. Then both scores are merged together to produce the final decision. In [14], an algorithm using a late fusion of dense LiDAR-based and image-based detections is presented. The authors apply region extraction and unary classification for each source separately. The fusion of the image and LiDAR detections is made by finding associations between the object candidates and fusing their bounding boxes. Instead of using two independent classifiers for LiDAR and image, we generate candidate regions employing solely Li-
DAR and classify them based on the visual information. The
algorithms described in [5, 18] are different from our approach
as they only use a one-layer laser, hence reducing the complex-
ity of the point cloud at the cost of reducing the precision of the
bounding boxes. In our framework, the laser scanner has 64
layers, which is important to produce high quality proposals.
Indeed, the measurements cover the entirety of the objects and
they provide precisely the geometry of each cluster. The advan-
tage of our method lies within the different processing steps of
the point cloud, which significantly improves the quality of the
clusters.

### 2.2 Visual Pedestrian Detection

A CNN is an artificial neural network that contains many con-
volutional layers. Those layers learn multiple filters, usually
of small size, to convolve the input. The flexibility of CNNs
allows them to be constantly improved with novel architecture
design. These novelties can modify the training time (resid-
ual network [11]), classification speed (region proposal net-
work [10, 16]), or performance (deeper or more complex net-
work). Due to its state-of-the-art performance, in this paper,
we use a residual network (ResNet) [11] for the detection. The
key contribution of the ResNet is to add an identity opera-
tion in the convolutional layers to connect the input and out-
put of each residual block and propagate only the difference
between the current block input and output. This difference
is used in the following layers, allowing it to learn complex
structures faster. An important factor contributing to the use of
neural networks in computer vision is called transfer learning:
this technique consists of using a CNN that has already been
trained on another database to considerably reduce the training
time [20]. This technique is employed in this work to fine-tune
the ResNet.

In order to evaluate the region proposal approach presented
in section 3, we compare it to a region proposal network (RPN)
used in Faster-RCNN [16]. RPNs are intended to reduce the
search space on an image by extracting regions of interest us-
ing a neural network. A region proposal network infers the
bounding boxes from the image itself prior to the classification
task. It should be noted that, in this study, we only focus on
showing the benefits of using a laser scanner to generate region
proposals and not on the whole detection framework.

## 3 Methodology

The LiDAR sensor renders a dense and accurate three-
dimensional point cloud as depicted in Figure 2a. Generat-
ing candidates for classification is performed by clustering this
point cloud. The number of clusters is then reduced in the val-
idation process. Afterward, clusters are projected on the image
space and gone through visual aspect correction to produce the
final candidates. The quality of region proposals generated by
the depth measurements is sensitive to the calibration and to
the processing of the three-dimensional point cloud. As the
density of the points can be high and have a negative impact on
the quality of clustering and computation time, we decided to
apply downsampling and to remove the points corresponding
to the ground.

### 3.1 Production of image proposals

**Data reduction**  Reducing the density of the LiDAR point
cloud improves the speed of the clustering without compromis-
ing efficiency. The density is a function of the distance from
sensor and follows a square rule:

$$density \propto distance^2$$

therefore the distribution of the points is not deterministic, that
is, it depends on the scene geometry. Downsampling is applied
as follows: a density reference is chosen and the distance axis
is then discretized. For all the points in one discretization step,
the data is reduced according to the density reference, if the
density is higher than the reference value. The resulting density
difference is illustrated in Figure 2b.

**Ground extraction**  The motivation behind extracting floor
points is to facilitate the clustering process. To perform ground
extraction, we assume that the lower points in the z-axis be-
long to the floor. We extract them by discretizing the floor (x
and y-axis) with a step given as parameter, and for each square
the lowest (z-axis) point is kept. Then, the other points in a
reasonably small distance from this lower reference are also
counted as floor points. The plane is found by computing poly-
nomial least-square fitting of degree two. This approach has
been chosen for its speed and simplicity and the outcome of
this process is visible in Figure 2c. The random sample consen-
sus (RANSAC) algorithm was also considered [6]. The latter is
iterative, hence the computation time is non-deterministic. Ad-
ditionally, the nature of the LiDAR sensor generates irregular density measurements: objects have a higher point density than the ground and, therefore, it can alter RANSAC performance.

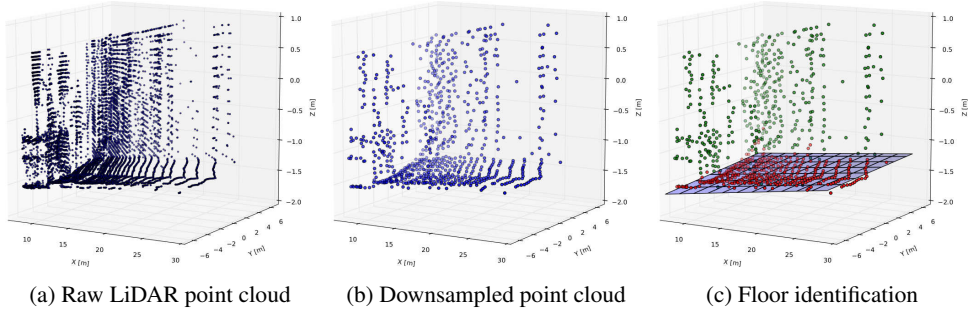

(a) Raw LiDAR point cloud  (b) Downsampled point cloud  (c) Floor identification

Figure 2: LiDAR data pre-processing

**Clustering** We require a simple and fast clustering algorithm that does not need any initialization. According to these criteria, we choose the "Density Based Spatial Clustering of Applications with Noise" (DBSCAN) algorithm [3]. It is a density-based algorithm designed on the concepts of *density-reachability* and *density-connection*:

1. *density-reachable*: a point $p$ is density reachable from a point $q$ if there is a chain of points $p_1, ..., p_n, p_1 = p, p_n = q$ such that $p_{i+1}$ is directly density-reachable from $p_i$. A point $p$ is directly density-reachable if the point $p$ is included in the area defined by a circle centered on $q$ of radius EPS.

2. *density-connected*: a point $p$ is density connected to a point $q$ if there is a point $o$ such that both, $p$ and $q$ are density-reachable from $o$.

The algorithm visits all points once and for each $p$ aggregates all *density-reachable* points according to the parameters EPS and MinPts. MinPts defines the minimum number of points that a cluster should contain, otherwise the group is considered as noise. EPS is a parameter that defines the maximum allowed distance between two *density-reachable* points. By projecting the clusters into the image space, we generate the candidates for detection (see Figure 3a).

**Validation, ratio and size adjustment** In order to generate more accurate candidate proposals, we make assumptions on the visual aspect of a pedestrian. A candidate is considered not valid if the width ($\Delta x$) of the cluster is greater than $0.1[m]$, the height ($\Delta y$) greater than $0.4[m]$ or the depth ($\Delta z$) lower than $1.2[m]$.

The shape of the candidates are then changed in two ways: the lower bound of the bounding box is adapted to match the ground floor and the aspect ratio of candidates is adjusted to a fixed value. This operation is useful to have a constant input size for classification and to avoid stretching effects when resizing. To select the best parameters for aspect ratio correction, we focus on the best precision possible for each parameter value. Results after this step are shown in Figure 3b.

## 3.2 Classification

The classification of the candidates is performed by a convolutional neural network. In this paper, we use the popular, high performing Residual Network (*ResNet*) [11]. The particularity of this network lies in the new architecture composed of residual blocks: they provide the advantage of a more stable training and a faster convergence [11]. The network has been pre-trained on the ImageNet [17] dataset. We have fine-tuned it for pedestrian detection on a reduced version of the KITTI training set that contains 3740 frames. During fine-tuning, we apply data augmentation by flipping the input images, hence doubling the amount of positive input samples. Optimization is performed by stochastic gradient descent with a learning rate value of $1e^{-4}$. The training took 2500 iterations with a batch size of 64.

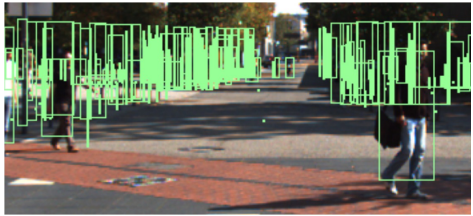

(a) Cluster proposal

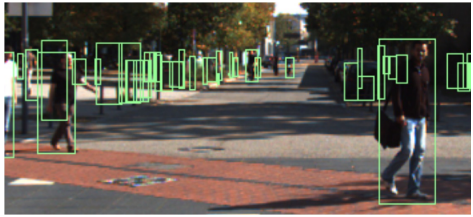

(b) Size and ratio corrections

Figure 3: Visualization of the region proposal

## 4 Experiments

In this section, we will first present the dataset and the evaluation metrics. Then the experimental results will be conveyed and discussed.

### 4.1 Dataset and evaluation

We have used the KITTI dataset [8] for the experiments. The particularities of this dataset regarding pedestrian detection are that some labels are highly occluded and the number of objects of small size is high. The y-axis size varies from 13 to 294 pixels for pedestrians. Moreover, the centering and alignment of the labels are not coherent through the images, and consequently introduce difficulties for the classifier to learn how to localize the candidates precisely.

The training set is composed of 7481 images and labels of the test set are not available. As a consequence, all the results reported in this work are computed on the validation set. Similar to [15], the provided training set is split into two subsets that are used as training set and validation set for our experiment: 3740 frames are used for training, and 3741 for validation.

The label matching criterion is an intersection over union (*IoU*) of 50%, $IoU > 0.5$, described in the PASCAL VOC challenge [4, 8].

$$IoU = \frac{area(B_p \cap B_{gt})}{area(B_p \cup B_{gt})}$$

With $B_p$ the detection bounding box and $B_{gt}$ the ground truth bounding box. Multiple detections of the same object are counted as false positives ($FP$).

### 4.2 Experimental results

The comparison of different region proposal schemes in terms of their effects on misses and recall rates are given in Table 1. In the table, *Clustering* refers to our proposed method based on LiDAR data. *Sliding window* refers to analyzing the image in a sliding window scheme. Faster R-CNN [16] uses its own RPN based on visual information. As can be observed the proposed approach provides fewer misses and higher recall rates. Especially, compared to a region proposal framework based on visual information, i.e. the one utilized at Faster R-CNN, the decrease of miss detection rate is significant. This indicates that LiDAR data can be utilized, in addition to visual information, to improve the performance of CNN-based pedestrian detection systems by reducing the miss detections and increasing the recall. Please note that, generally, the recall rate is very sensitive to the aspect ratio of the proposals. For performance comparison, we fixed the aspect ratio at the value which minimizes the number of missed labels. Additionally, the parameters of the different approaches can impact significantly the recall value. For example, sliding window can fail to overlap adequately two labels that are close to each other. Our approach covers two close objects more efficiently when they are clustered separately. Although the Faster R-CNN is a generic object detection framework, we carefully adapted the aspect ratio of the resulting proposal output to have a representative comparison.

Figure 4 plots the recall rates with respect to different *IoU* values. Similar to the findings in Table 1, LiDAR clustering achieves higher recall rates also at different *IoU* values. Moreover, compared to the sliding window scheme, it reduces the number of regions from 4009 down to 307 as presented in Table 2. The computational overhead of determining candidate regions is negligible as can be seen in the last column of Table 2. Thus, the proposed approach reduces the amount of computation significantly with respect to the sliding window scheme. Compared to the visual information based region extraction as in Faster R-CNN, employing LiDAR data requires more processing time. However, considering the detection times, listed in the second column of Table 2, this difference is also negligible. The detection time refers to the inference time when the regions are fed one by one to the classification network.

| Region extraction | Missed labels (FN) | Max recall |
|---|---|---|
| Clustering | 180 | 0.92 |
| Sliding window | 219 | 0.90 |
| Faster R-CNN [16] | 601 | 0.73 |

Table 1: Number of labels that are missed (FN) and maximum recall possible on the validation set with an *IoU* of 0.5

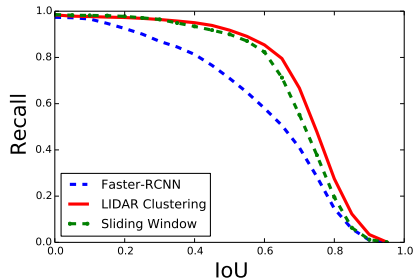

Figure 4: Comparison of recall rates with respect to different *IoU* values

| Region extraction | Number of regions | Detection time | ROI time |
|---|---|---|---|
| Clustering | 307 | 21 [s] | 0.5 [s] |
| Sliding window | 4009 | 219 [s] | - [s] |
| Faster R-CNN [16] | 300 | 21 [s] | 0.3 [s] |

Table 2: Comparison of different region extraction methods

The final performance of pedestrian detection in real use-case scenario strongly depends on the tuning of parameters, bounding box adaptations, and non-maximum suppression function. Indeed, the max recall is bounded by the region proposal approach as shown in the previous experiment, and the precision highly depends on the non-maximum suppression employed after detection.

Table 3 shows a clear difference when using cluster candidates compared to using sliding window. As expected, the act of reducing the number of candidates to classify impacts the *precision* by decreasing the number of false positives. We observe an absolute increase of around 20% on medium difficulty average precision. The *recall* is impacted as well by the decrease of the number of false negatives. Please again note that the purpose of the study is to show the benefits of employing LiDAR data to improve region proposals. Therefore, we combined the proposed approach with a generic state-of-the-art object classification framework, namely, *ResNet*. Building an elaborate and optimized CNN-based pedestrian detection system is beyond the scope of the paper. However, the presented ideas, i.e. exploiting LiDAR information for improved region proposals, can also be incorporated to the state-of-the-art vision-only pedestrian detection approaches.

| Detection | AP easy | AP medium | AP hard |
|---|---|---|---|
| ResNet, sliding window | 35.8 % | 34.3 % | 31.2 % |
| ResNet, clustering | 56.4 % | 54.5 % | 50.4 % |

Table 3: Average precision (AP) for the different detection schemes

## 5 Conclusion

In this paper, we presented a novel region proposal framework based on depth measurements from the LiDAR. The experimental results showed the range of performance gain using our region proposal approach. It provides reduction in the image search space, the amount of miss detections and increase in recall. An advantage of our region proposal resides in the fact that it can be applied prior to any detection framework. This research can therefore be continued by extending the results with more efficient and deeper networks trained for pedestrian detection. In summary, LiDAR data can provide complementary information to the visual information and can be utilized to improve the CNN-based pedestrian detection approaches further.

## Acknowledgments

This material is based upon work supported by the international chair DriveForAll [1].

---
[1] http://driveforall.com/, accessed July 2017

7